\documentclass{article}

\PassOptionsToPackage{numbers, compress}{natbib}
\usepackage{neurips_data_2022}

\usepackage[utf8]{inputenc} % allow utf-8 input
\usepackage[T1]{fontenc}    % use 8-bit T1 fonts
\usepackage{hyperref}       % hyperlinks
\usepackage{url}            % simple URL typesetting
\usepackage{booktabs}       % professional-quality tables
\usepackage{amsfonts}       % blackboard math symbols
\usepackage{nicefrac}       % compact symbols for 1/2, etc.
\usepackage{microtype}      % microtypography
\usepackage{xcolor}         % colors
\usepackage{amsmath}
\usepackage{authblk}
\usepackage{graphicx}
\usepackage{multirow}
\usepackage{makecell}
\usepackage[symbol]{footmisc}
\newcommand{\eg}{\textit{e}.\textit{g}. }
\newcommand*\samethanks[1][\value{footnote}]{\footnotemark[#1]}

\title{AnoShift: A Distribution Shift Benchmark for Unsupervised Anomaly Detection}

\author[,,1]{Marius Dragoi\thanks{Equal contribution. $^1$Bitdefender Theoretical Research Team: \url{https://bit-ml.github.io/}}}
\author[,,1,2]{Elena Burceanu\samethanks}
\author[,,1,3]{Emanuela Haller\samethanks}
\author[1]{Andrei Manolache}
\author[1]{Florin Brad}

\affil[1]{Bitdefender, Romania}
\affil[2]{University of Bucharest}
\affil[3]{Politehnica University of Bucharest}

\begin{document}

\maketitle

\begin{abstract}
Analyzing the distribution shift of data is a growing research direction in nowadays Machine Learning, leading to emerging new benchmarks that focus on providing a suitable scenario for studying the generalization properties of ML models. The existing benchmarks are focused on supervised learning, and to the best of our knowledge, there is none for unsupervised learning. Therefore, we introduce an unsupervised anomaly detection benchmark with data that shifts over time, built over Kyoto-2006\texttt{+}, a traffic dataset for network intrusion detection. This kind of data meets the premise of shifting the input distribution: it covers a large time span ($10$ years), with naturally occurring changes over time (\eg users modifying their behavior patterns, and software updates). We first highlight the non-stationary nature of the data, using a basic per-feature analysis, t-SNE, and an Optimal Transport approach for measuring the overall distribution distances between years. Next, we propose \textbf{AnoShift}, a protocol splitting the data in IID, NEAR, and FAR testing splits. We validate the performance degradation over time with diverse models (MLM to classical Isolation Forest). Finally, we show that by acknowledging the distribution shift problem and properly addressing it, the performance can be improved compared to the classical IID training (by up to $3\%$, on average). Dataset and code are available at \url{https://github.com/bit-ml/AnoShift/}.

\end{abstract}

% keywords:  non-stationary, distribution shift, anomaly, unsupervised, network intrusion detection, network traffic, natural distribution shifts

\section{Introduction}

Analyzing and developing Machine Learning algorithms under gradual distribution shifts is a problem of high interest in the research community. 
There is a growing enthusiasm for building benchmarks over existing or new datasets~\cite{clear, mind_the_gap, shift_auto_drive, intel_cl_benck, wilds}, that formulate a setup for isolating the shifting aspect and create a better ground for this research field. A better understanding of the distribution shift problem might lead to findings of underlying fundamental aspects, shedding new light on robustness and generalization problems. We argue that the distribution shift occurs naturally and gradually in a continuous data stream (\eg monitoring network traffic), allowing an in-depth analysis of the problem. On the other side, artificially generated scenarios usually exhibit sudden changes that do not simulate the natural shift problem. Yet, the annotation process for streaming data is quite difficult and expensive, considering the massive amount of data. 

%Analyzing and developing Machine Learning algorithms for data under gradual distribution shift is currently a problem of high interest in the ML research community. This started an effervescence of newly introduced benchmarks over existing or new datasets~\cite{clear, mind_the_gap, shift_auto_drive, intel_cl_benck, wilds} that clearly formulate setups to isolate this shifting aspect and create a better scene for researching this direction. The motivation behind researching the capabilities of the models to adapt to distribution shift in data, is a high aspiration in the Machine Learning field, and might lead to findings of underling fundamental aspects that bring robustness and generalization to ML algorithms.

%We argue that in continuous streams of data, coming from dynamic systems like the internet traffic from a network, this distribution shift occurs naturally and smoothly (gradually) and it can be better highlighted and studied. This contrasts with other cases with more rigid, artificially built setups that exhibits sudden changes. But one main disadvantage of the streaming data, is that it usually comes in large uninterrupted volumes, making the labeling task quite difficult and expensive, and also rare. 

From a practical point of view, continuous IT infrastructure monitoring has become essential for computer security and resilience. Recent anomaly detection and intrusion detection systems (IDS) obtain strong results on specific datasets but drastically fail in real world scenarios \cite{nids_generalization}. Our proposed benchmark relies on the network traffic data extracted from the popular Kyoto-2006\texttt{+}, tackling the problem of unsupervised intrusion detection. Our experimental analysis of the considered dataset proved that there is a natural change over the $10$ years during which the data was collected. The shift is noticeable both over the input distribution and considering the performance of several anomaly detection systems. Several reasons behind the observed shift are: users leaving or coming to the network, per user interest changes leading to network interaction changes, updates to the software versions, patching old vulnerabilities but revealing new attack vectors for intruders.
%From the practical perspective, continuous monitoring of IT infrastructure has become an important process of computer security and resilience. One of the key tools of this process is log-based anomaly detection. Recent anomaly detection and intrusion detection systems (IDS) are able to obtain very strong results on several datasets, saturating them very quickly. But, in real world scenarios, such systems fail drastically\cite{nids_generalization}.

%We consider that their performance is heavily influenced not only by the data itself, but by several experimental setup decisions, which may lead to performance inconsistencies. Therefore, we argue that the common assumptions that the data samples are identical and independently distributed do not take into account the distribution shifts over time, leading to inflated scores on the test set. In our approach, after investigation the datasets landscape for this unsupervised intrusion detection task in network traffic, modeled as anomaly detection, we settled to the Kyoto-2006\texttt{+} dataset. We experimentally prove by analysing its input distribution, but also the performance of several models, that it is suitable for noticing distribution shifts, as its data naturally change over $10$ years period. Several reasons behind this are: users leaving or coming to the network, per user interest changes leading to network interaction changes, updates to the software versions, patching old vulnerabilities, but revealing new attack vectors for intruders.

\begin{figure}[t]
    \begin{center}
        \includegraphics[width=0.9\textwidth]{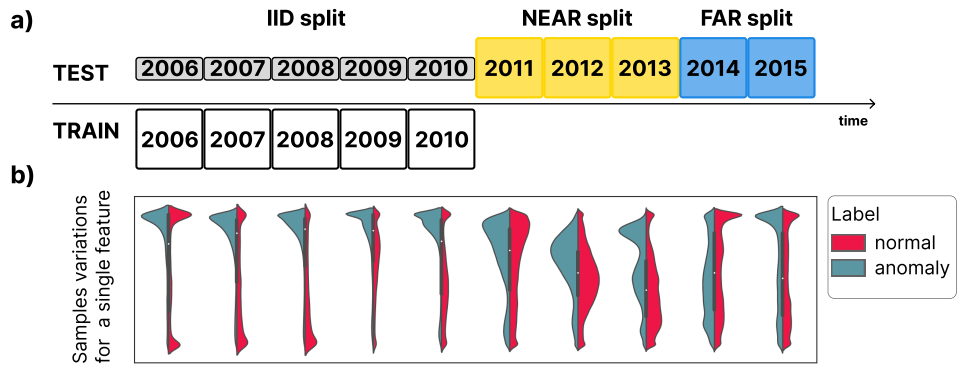}
    \end{center}
    \caption{\textbf{a)} The proposed \textbf{AnoShift} splits over Kyoto-2006\texttt{+} dataset. The \texttt{IID} (gray) testing split comes from the same temporal span as the \texttt{TRAIN} set (white), while \texttt{NEAR} (yellow) and \texttt{FAR} (blue) splits are from different time spans, with \texttt{NEAR} being closer to the training set than \texttt{FAR}. \textbf{b)} To highlight the utility of the proposed chronological protocol, we exemplify the continuous evolution of data, illustrating the distributions of normal and anomaly samples over the considered 10 years. We exemplify the evolution of the percent of recent connections that have the same source and destination IP addresses as the current connection (\href{http://www.takakura.com/Kyoto_data/BenchmarkData-Description-New.pdf}{feature 9 - Dst host srv count}).}
    %The proposed Kyoto-2006\texttt{+} split for emphasizing data distribution shift. \texttt{NEAR} (yellow) split is closer to the training data, while the \texttt{FAR} (blue) splits is farther. We exemplify with green the distribution for a single feature. Notice how it continuously changes over time.\answerTODO{alt feature. care sa arate schimbare catre FAR mai mare}}
    \label{fig:anoshift_splits_feat_distrib}
    \vspace{-1em}
\end{figure}

To better assess the models' capabilities, we introduce a chronology based evaluation protocol, distinctly evaluating performance on data splits ((\texttt{IID}, \texttt{NEAR} and \texttt{FAR} - Fig.~\ref{fig:anoshift_splits_feat_distrib})) with different temporal distances towards the training set (\texttt{TRAIN} -Fig.~\ref{fig:anoshift_splits_feat_distrib}). We observe that the performance of anomaly detection models consistently degrades when tested on data from longer time horizons. Moreover, we prove that a basic distillation technique overcomes a classic IID training under gradual data shifts, proving that the awareness of the shift problem might lead to better solving the task.

%To better assess the models' capabilities, we propose a chronology based evaluation protocol over Kyoto-2006\texttt{+} dataset, in which we distinctly evaluate anomalies that happen close to the training points (\texttt{NEAR} split) versus anomalies that are more distant in the future (\texttt{FAR} split), as illustrated in Fig.~\ref{fig:anoshift_splits_feat_distrib}. We observe that the test performance of anomaly detectors for network traffic consistently degrades with longer time horizons, for several tested models. Moreover, we show that the awareness of the shift problem might lead towards better solving the task. Taking advantage and implementing a basic distillation technique as opposed to the IID classic training improved the overall results.

Summarized, our \textbf{main contributions} are the following:
\begin{itemize}
%\itemsep0em
    %\vspace{-0.1cm}
    \item We analyzed a large and commonly used dataset for the unsupervised anomaly detection task in network traffic (Kyoto-2006\text{+}) and demonstrated that it is affected by distribution shifts. The per-feature distributions and t-SNE show multiple changes over the years, and the Optimal Transport Dataset Distance~ gave us an estimate of its magnitude.
    %\vspace{-0.1cm}
    \item We propose a chronology-based benchmark, which focuses on splitting the test data based on its temporal distance to the training set, introducing three testing splits: \texttt{IID}, \texttt{NEAR}, \texttt{FAR} (Fig.~\ref{fig:anoshift_splits_feat_distrib}). This testing scenario proves to capture the in-time performance degradation of anomaly detection methods for classical to masked language models. This benchmark aims to enable a better estimate of the model's performance, closer to the real world performance. 
    %\vspace{-0.1cm}
    \item We prove that properly acknowledging the distribution shift may lead to better anomaly detection models than classical IID training. When facing distribution shifts, a basic distillation technique positively impacts the performance by up to $3\%$ on average.
\end{itemize}

%Summarized, our \textbf{main contributions} are the following:
%\begin{itemize}
%    \item We analysed a large and commonly used dataset for the unsupervised anomaly detection task in network traffic (Kyoto-2006\text{+}) and demonstrate that it clearly suffers from distribution shift. The per-feature distributions and t~SNE show that there are multiple changes over the years, and the Optimal Transport Dataset Distance give us an estimate of its magnitude.
%    \item We propose a chronology-based benchmark for evaluating anomaly detection in network traffic, which focuses on splitting the test data based on its temporal distance to the training set: \texttt{NEAR} and \texttt{FAR} splits (close, respectively farther apart). In general, we found that the performance of the model degrades over time, for classical models but also for those trained as masked language models. This benchmark aims to enable the validation of better models, in a scenario that is closer to the real world.
%    \item We further show that by acknowledging the distribution shift problem and by properly addressing it, the performance can be improved compared to the classical iid training. When the distribution changes, a basic distillation technique impacts the performance in a positive way (by over 3\% on average).
%\end{itemize}

\section{Related work}

% Online Learning/Continuous Learning

\textbf{Relation to benchmarks targeting distribution shift} Recently, there has been an increased amount of effort and focus in this direction, with several benchmarks emerging. They emphasize the non-stationary nature of the data, with various underlying reasoning. The most common approach is to search for gaps in the input data distribution that appear with time~\cite{clear, mind_the_gap}, taking into perspective that the world is continuously evolving; therefore, the data acquired continuously from it should exhibit the same behavior. Our work aligns with this perspective by working with traffic logs from a large university network over $10$ years. In~\cite{clear}, the authors focus on how the appearance of basic objects changes from year to year, while~\cite{mind_the_gap} emphasizes the seasonal patterns that appear in news language (\eg elections, hurricanes). A second axis exploited for noticing shifts in data is the spatial one. In~\cite{intel_cl_benck}, geolocalization is used in conjunction with the time for guiding the shift. In~\cite{wilds}, the gap is based on higher level characteristics, like x-ray data from different hospitals, but also on geolocalization. In searching for the autonomous driving robustness, a more complex variation is provided in ~\cite{shift_auto_drive} following the weather, time of day, and congestion levels. Nevertheless, all works analyze the distribution shift for supervised tasks, focusing on NLP or Computer Vision. In~\cite{mind_the_gap}, the authors monitor the evolution of the perplexity metric, with models learned in a self-supervised manner as a masked language model. They emphasize the need to link the shift analysis to a downstream task, several supervised ones in their case. Differently, \textbf{AnoShift}, our benchmark proposal, tackles an unsupervised anomaly detection task under non-stationary data.

\textbf{Relation to traffic anomalies} Models tackling Network Intrusion Detection are covered by lots of surveys~\cite{nids-survey, nids_survey_3, nids_survey_4}, structured around dataset variations, anomaly types, and methods variation. A fair amount of the approaches are supervised~\cite{compare_nids}, based on tree classifiers~\cite{tree}, modeling the task as a binary or multi-class anomaly (intrusion) classification. But we are interested here in the unsupervised setup~\cite{ensemble_autoenc_nids}. Usually, the best models are quite simple, most of them are shallow~\cite{shallow_deep_nids}, based on OC-SVM~\cite{ocsvm} or Isolation Forest~\cite{isoforest}, or very small neural nets~\cite{ensemble_autoenc_nids}. Several solutions introduce deep learning approaches for intrusion detection~\cite{ DL_anom_survey}, transforming the data into images~\cite{nids_images_svm}, or modeling the problem using GNNs~\cite{gnn_nids}. 

An important problem we identified in this area is that the datasets used for the task are easily saturated, mainly because they either lack variety (\eg simulated traffic patterns for anomalies) or have a very few annotated anomalies, or are small-scale, covering only several days~\cite{kddcup, NSL-KDD, cic_ids_2017, netflow_v1_iot, netflow_v2_iot, unsw_nb15, lanl, zyell_nctu, compare_nids}. In contrast, Kyoto-2006\texttt{+}~\cite{kyoto2006} spans over $10$ years (2006-2016), containing continuous natural traffic logs from a large university network, within a sub-net of honeypots. Most of those datasets cover basic networks, but there are some oriented towards IOT traffic~\cite{netflow_v2_iot}, or even to the autonomous driving field, Internet of Vehicles~\cite{tree}. But another reason for saturation, is the IID training setup, as we will show in this work. These generalization problems are very acute, leading to weak performances for those algorithms when applied on real world data, or on a new dataset~\cite{nids_generalization}. With \textbf{AnoShift}, we highlight the IID training problem, by proposing a different training and evaluation setup based on temporal distances, closer to a realistic case.

\section{Chronological protocol}\label{sec:chronological_protocol}
%\begin{itemize}
%    \item \answerTODO general description of the protocol - how you organize data 
%    \item \answerTODO ref to the considered task - NIDS 
%\end{itemize}

%\paragraph{IID, NEAR, FAR splits} To emphasize the decrease in performance of models when dealing with temporarily distant data from the training set, we merge multiple yearly splits, on the Kyoto-2006\texttt{+}. We create two separate test sets, one close to the training data: NEAR and one farther apart: FAR.  We expect models to exhibit a better performance on \texttt{NEAR} compared to \texttt{FAR}, which we experimentally prove in Tab.~\ref{tab:far_near}. We specify bellow the \texttt{IID}, \texttt{NEAR} and \texttt{FAR }splits for each dataset, in its subsection and in Fig.~\ref{fig:counts}.

We introduce a chronological protocol for building train and testing splits that can highlight the temporal evolution of data. Taking into consideration the timestamps of our data, we propose to build a training split (\texttt{TRAIN}) along with three different testing splits (\texttt{IID}, \texttt{NEAR} and \texttt{FAR}), comprising multiple years of data (Fig.~\ref{fig:anoshift_splits_feat_distrib}a)). The \texttt{TRAIN} and \texttt{IID} splits are extracted from the first period of time, and the \texttt{IID} tests should highlight the expected performance when there is no distribution shift between train and test. The \texttt{NEAR} and \texttt{FAR} splits are each extracted from different periods of time, where \texttt{NEAR} is closer to the training data and \texttt{FAR} is farther away.  We expect standard models to exhibit better performance on \texttt{NEAR} compared to \texttt{FAR}, which we experimentally prove in Sec.~\ref{sec:distrib-shift_models}. 
Our proposed benchmark will provide a better estimate of the expected performance when the model is deployed in the wild and exposed to the inevitable distribution shift of the data. To the best of our knowledge, \textbf{AnoShift} is the first to provide a proper scenario for studying the generalization capabilities of unsupervised learning models for anomaly detection. 

Our work revolves around Network Intrusion Detection Systems (NIDS), tackling the problem of distribution shifts that naturally appear in network traffic data. We work over the popular Kyoto-2006\texttt{+} dataset (Sec.~\ref{sec:kyoto}), which was collected over ten years, providing us with enough data to capture the temporal evolution. Starting from Kyoto-2006\texttt{+}, we introduce our \textbf{AnoShift} Benchmark (Sec.~\ref{sec:anoshift}) that proposes one training and three testing splits, which highlight the difficulty of dealing with data temporarily distant from the training set. 

\subsection{Kyoto-2006+}\label{sec:kyoto}
%\begin{itemize}
%    \item \answerTODO general description + example 
%\end{itemize}

Kyoto-2006\texttt{+}~\cite{kyoto2006} is a reference dataset for Anomaly Detection over network traffic data~\cite{ring2019survey}. It is built on $10$ years of real traffic data (Nov. 2006 - Dec. 2015), captured in a system of $348$ honeypots in $5$ sub-networks inside the Kyoto University. Briefly, a honeypot is a real or virtual machine simulating a regular computer (having an OS and multiple services running on it). Its purpose is to deceive an attacker into taking advantage of the vulnerabilities present on the honeypot machine (\eg software not updated). A honeypot does not request any connection on its own. So in such a scenario, almost all traffic coming to a honeypot machine is unsolicited and therefore considered malicious. By design, this kind of dataset has a large percent of anomalies (89.5\% anomalies in Kyoto-2006\texttt{+}) compared to other anomaly detection datasets. The 14 conventional features of the dataset include 2 categorical ones like connection service type or flag of the connection and 12 numerical like the connection duration or the number of source bytes. We put more details about Kyoto-2006\texttt{+} in Appendix~\ref{appendix}. This dataset is spread across a very large period of time, and it contains exclusively real-world traffic events without simulated events.

\subsection{AnoShift benchmark}\label{sec:anoshift}
    %\begin{itemize}
    %    \item \answerTODO how do we split the dataset -> plots with nr samples \& figure with train test splits
    %    \item \answerTODO justification with simple plots - per feature distribution changes - link to in depth analysis from sec 4
    %\end{itemize}
    
\begin{figure}[t]
    \begin{center}
        \includegraphics[width=0.99\textwidth]{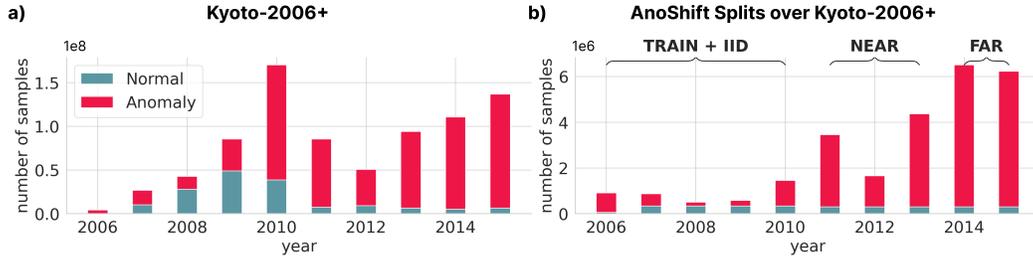}
    \end{center}
    \caption{\textbf{a)} Yearly splits of the network traffic data from Kyoto-2006+ dataset, highlighting the proportion of normal and anomaly samples. \textbf{b)} Proposed train and test splits in our \textbf{AnoShift} benchmark. Considering that \texttt{TRAIN} and \texttt{IID} splits are sampled from the same time span, we have jointly represented them. Note that while for \texttt{TRAIN} , \texttt{NEAR} and \texttt{FAR} we extract the same number of normal samples per year, the \texttt{IID} split contains 10 times less normal samples. The anomaly samples are extracted such that we maintain the normal vs. anomaly proportion of the original data.}
    \label{fig:anoshift_n_samples}
    \vspace{-1em}
\end{figure}

    To keep the natural distribution shift of the network traffic data, we sample a fixed number of normal samples per year (\#months$\times$25k for \texttt{TRAIN}, \texttt{NEAR}, and \texttt{FAR} and \#months$\times$2.5k for \texttt{IID}). The number of anomalous samples is chosen such that we maintain the proportion of normal vs. anomaly samples from the original yearly split. We illustrate this process in Fig.~\ref{fig:anoshift_n_samples}. In Fig.~\ref{fig:anoshift_splits_feat_distrib} b) we illustrate the continuous evolution of the data features over the considered 10 years, comparing the distribution for one feature (\href{http://www.takakura.com/Kyoto_data/BenchmarkData-Description-New.pdf}{feature 9 - Dst host srv count}), over the years. Such behavior can be observed for the majority of features, a fact highlighted by our in-depth analysis from Sec.~\ref{sec:distrib-shift_metrics}. The \texttt{TRAIN} and \texttt{IID} samples are collected from $[2006-2010]$, while the \texttt{NEAR} and \texttt{FAR} splits consist of $[2011-2013]$ and $[2014-2015]$ intervals. The protocol is illustrated in Fig.~\ref{fig:anoshift_splits_feat_distrib} and Fig.~\ref{fig:anoshift_n_samples}.

    \subsubsection{Experimental setup}
    \label{sec:exp_setup}
    %\begin{itemize}
    %    \item \answerTODO preprocess 
    %    \item \answerTODO eval 
    %\end{itemize}
    
    \paragraph{Preprocess network traffic data} We use the 14 conventional features from the new version of the Kyoto dataset [2006-2015] and convert the numerical features to categorical values by using an exponentially-scaled binning method between 0 and the maximum value of each feature, such that the bins have a higher density for smaller numerical values and are increasingly wider towards larger values. We used a basis of 1.1, which resulted in 233 bins, where the width of the $i$th bin is given by: $ bin_i = [1.1^{i} - 1, 1.1^{i+1} -1]$. We keep the original percentage features, which are discretized in 100 values. Therefore, our preprocessing results in a fixed vocabulary size, which allows language modeling, and each possible token is known apriori. See in Fig.~\ref{fig:kyoto_instance} a preprocessed sample. Our processing of the original dataset does not pose any privacy concerns since it does not contain any sensitive information, such as IP address. However, data binning constitutes another potential limitation in our work.
    
\begin{figure}[t]
    \begin{center}
        \includegraphics[width=1.0\textwidth]{images/kyoto_sample.png}
    \end{center}
    \caption{Examples of preprocessed Kyoto-2006\texttt{+} instances. See Appendix~\ref{appendix} for details.}
    \label{fig:kyoto_instance}
    \vspace{-1em}
\end{figure}

    \textbf{Metrics for anomaly detection} To analyze the performance of various models on our proposed benchmark, we use the labels (normal and anomaly) provided by the Kyoto-2006\texttt{+} dataset. As we deal with imbalanced sets, we study the performance of both ROC-AUC and PR-AUC. The PR-AUC is computed for both inliers and outliers to cope with positive or negative minority situations (note that for a random classifier, PR-AUC for a specific class is close to 0. We report the \texttt{IID}, \texttt{NEAR} and \texttt{FAR} performances as the arithmetic mean of performances over their associated yearly splits.
    
    %Since we discuss datasets splits that have labels and are highly unbalanced, ROC-AUC is the fair metric to use for comparing the performance in different scenarios. For better understanding the problems with different splits and models, we also report the Precision-Recall-AUC both for inliers and outliers in the appendix \answerTODO{}, but only for analysis, since this metric is highly unstable when the data proportion changes. Please note that for a random classifier, PR-AUC for a specific class is close to 0.

\section{Distribution shift analysis}\label{sec:distrib_shift_analysis}

    We perform an in-depth analysis of the proposed benchmark from three points of view. \textbf{First}, we study the inherent non-stationarity of the considered data, highlighting the natural shift between the years, considering both simple, per feature metrics and more complex metrics between distributions (Sec.~\ref{sec:distrib-shift_metrics}). \textbf{Second}, we analyze various anomaly detection models, highlighting the performance decrease when dealing with testing data that is temporarily distant from the training set (Sec.~\ref{sec:distrib-shift_models}). \textbf{Third}, we discuss the importance of acknowledging the data shift and emphasize the positive impact of a basic distillation technique over the standard IID approach (Sec.~\ref{sec:ack_shift}). We add supplementary discussions on the method in Appendix~\ref{appendix:discussions}.
    
    We run our experiments on an internal cluster with multiple GPU types: GTX 1080 Ti, GTX Titan X, RTX 2080 Ti, RTX Titan. We estimate that we need 3 days to reproduce the experiments on 1 GPU. The CPU training for OC-SVMs, IsolationForest, and LOF benchmarks takes ~3 days.

\subsection{Inherent non-stationarity}\label{sec:distrib-shift_metrics}
%\begin{itemize}
%    \item \answerTODO Jeffrey \& OTDD \& TSNE 
%\end{itemize}

\begin{figure}[t]
    \begin{center}
        \includegraphics[width=0.9\textwidth]{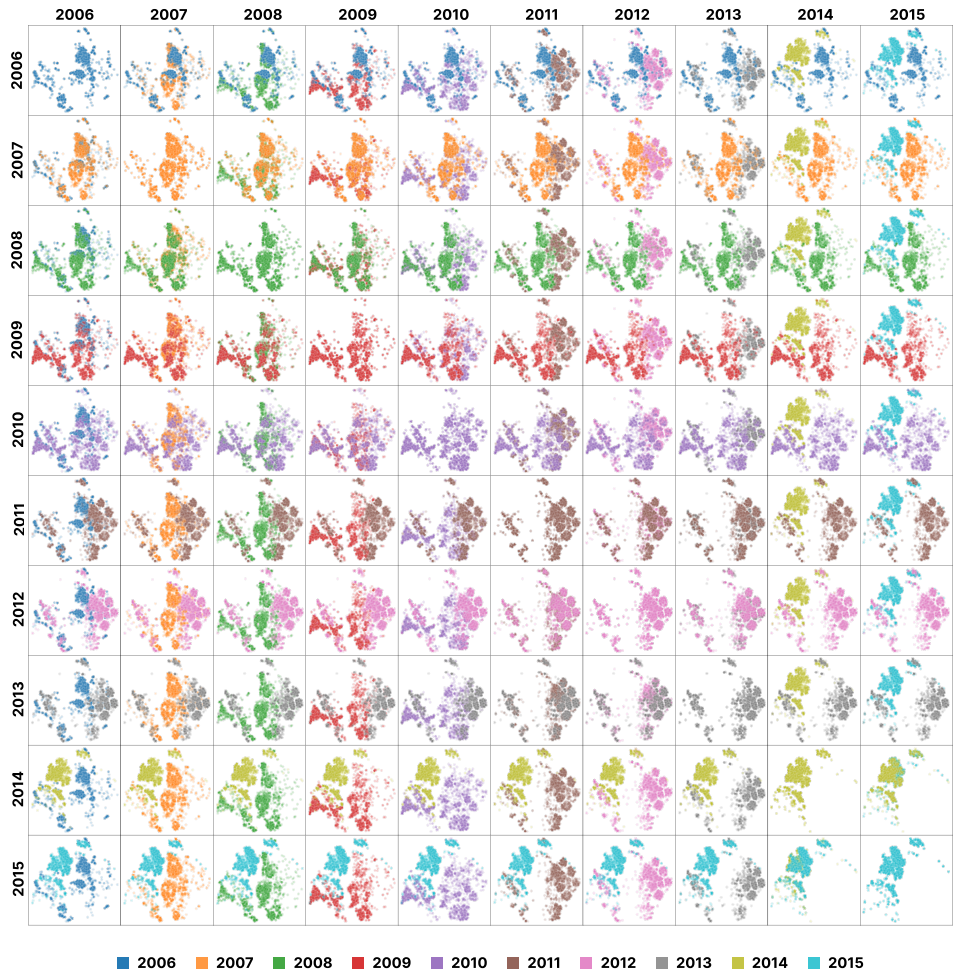}
    \end{center}
    \caption{Comparison between yearly splits using t-SNE visualization. We observe that the discrepancy between point clouds increases with the temporal distance between splits, colors becoming more separated over time. The analysis is performed considering 2k randomly sampled points per split. Follow the 2007 row: see the orange cluster on top of clusters associated to the other years. It is very similar to its neighbours 2006-2008, and the similarity diminishes in time (see 2015).}
    \label{fig:tsne_all}
    \vspace{-1em}
\end{figure}

    \textbf{Visualization of the data shifts} For a visual interpretation of the yearly shift, we have considered the unsupervised t-SNE~\cite{tsne} to illustrate our high dimensional data structure. In Fig.~\ref{fig:tsne_all} we present the comparison between pairs of yearly splits using t-SNE. We observe that the point clouds belonging to two different splits are getting further away from each other once we increase the temporal gap between the considered splits. This confirms our intuition that the analyzed network traffic data is continuously shifting in time and emphasizes the need for a benchmark as \textbf{AnoShift} that can efficiently test the robustness of models under this inherent non-stationarity of natural data.
    
    %\begin{figure}[t]
    %    \begin{center}
    %        \includegraphics[width=0.7\textwidth]{images_/tsne_pairs.png}
    %    \end{center}
    %    \caption{Comparison between consecutive pairs of yearly splits using t-SNE visualization. We observe that the discrepancy between point clouds increases with the temporal distance between splits. The analysis is performed considering 2k randomly sampled points from each split. (Best viewed in color)}
    %    \label{fig:tsne_pairs}
    %\end{figure}
    
\begin{figure}[t]
        \begin{center}
            \includegraphics[width=1\textwidth]{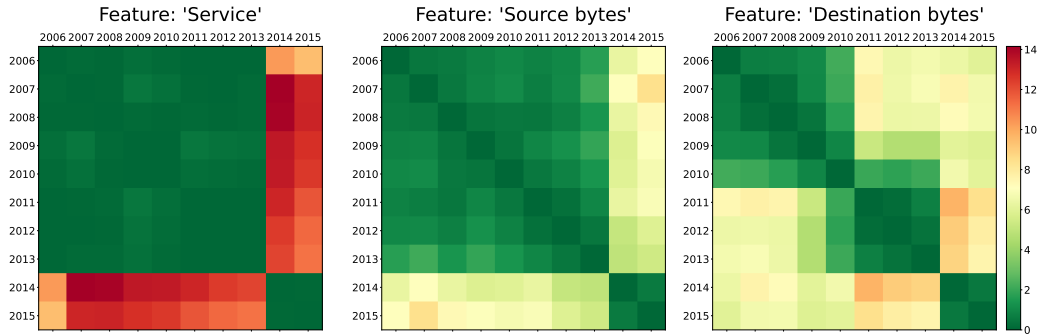}
        \end{center}
        \caption{Jeffreys divergence between Kyoto years. First two images represent features with a large 2014-2015 distance. The $3^{rd}$ one is for
        a feature with significant difference between the histograms across years. Note that it is difficult to predict the performance of the method on a new split, only based on those per feature distances between distributions.}
        \label{fig:jeff_kyoto}
    \end{figure}

    \textbf{Per-feature shift} We further analyze whether the dataset's statistics at the feature level are changing from one year to another. Recall that we have 2 categorical features and 12 numerical ones. We extract the normalized histogram per year for each feature and compute the Jeffreys divergence~\cite{jeff_divergence} between those histograms. The Jeffreys divergence is a commonly used symmetrization for Kullback-Leibler divergence~\cite{kl_divergence}: $KL(p, q) + KL(q, p)$, and it is proven to be both symmetric and non-negative. We highlight that such an analysis can only illustrate simple scenarios, studying the distribution change from the perspective of single feature changes. With all the considered baselines from Sec.\ref{sec:distrib-shift_models}, we have observed a significant decrease in performance for the years 2014 and 2015, leading to the intuition that this subset may have substantial differences from the others. Consequently, in Fig.~\ref{fig:jeff_kyoto}, we illustrate the Jeffreys divergence for two features that we find to have a large 2014-2015 distance, but also for a third one that has significant high values in the distance map on other years than the two.
    
\begin{figure}[t]
        \begin{center}
            \includegraphics[width=1\textwidth]{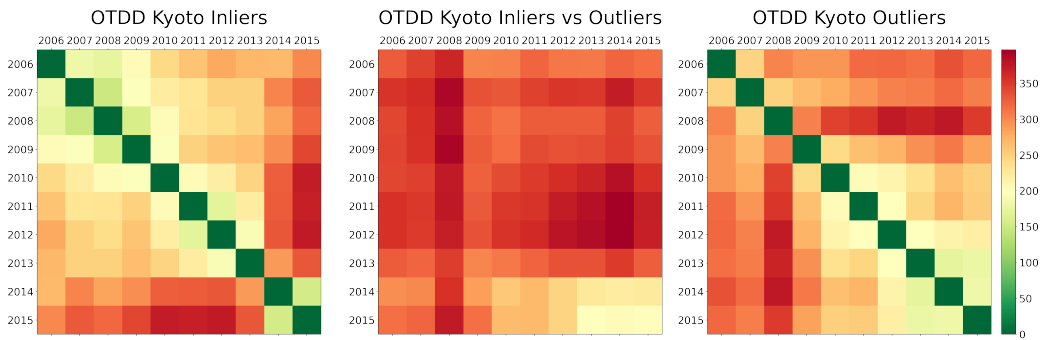}
        \end{center}
        \caption{Optimal Transport Dataset Distance for Kyoto. See distances between inliers (first), inliers and outliers (second), and outliers (third). The distances from inliers generally increase as you move further from the diagonal, showing large distances between \texttt{TRAIN} and \texttt{FAR} data. Moreover, notice in second image how outliers in the \texttt{FAR} splits are quite similar with inliers from \texttt{TRAIN}, also explaining the abrupt performance drop on farther data (Tab.~\ref{tab:far_near}).}
        \label{fig:otdd_kyoto}
        \vspace{-1em}
    \end{figure}

    \textbf{General shift} We next explore the distribution differences between dataset splits over time by using the Optimal Transport Dataset Distance method (OTDD)~\cite{otdd}. OTDD relies on optimal transport, a geometric method for computing distances between probability distributions for comparing datasets. This analysis shows how the splits move away from each other over time (see Fig.~\ref{fig:otdd_kyoto}). Compared with the per feature approach, this method allows us to gain a better intuition for the performance on a new split, giving us a single distance based on all features. We observe how the inliers (first image) nicely distances in OTDD value, directly correlated with the distance in time. As for the outliers (third image), it is noticeable that they are quite different between the splits of the first years. We notice that the distances between inliers and outliers (in the middle) show that \texttt{FAR} years' outliers are similar to \texttt{TRAIN} years' inliers, an observation that we empirically confirm in Tab.~\ref{tab:far_near}, where all models suffer from a steep descent in performance (bellow random). We run the method with the default parameters for DatasetDistance, over the standardized input of Kyoto, with one-hot encoded categorical variables, $3$ times, with a randomly sampled 5k entries per year.

\subsection{Impact on IID models}\label{sec:distrib-shift_models}
%\begin{itemize}
%    \item \answerTODO description of models - per category 
%    \item \answerTODO general eval + per month eval + impure train data (your splits affect the proportion \& purely unsup split => how bad?)
%\end{itemize}
    
    We introduce the \textbf{AnoShift} benchmark to understand better the impact of data shifts that naturally appear over time on the performance of anomaly detection models. We hope that the proposed splits will push forward the research in this direction and help build more robust models that can deal with mild to severe distribution changes between test and training sets. In this context, in the current section, we will study the performance degradation of various anomaly detection approaches, from \texttt{IID} to \texttt{NEAR} and \texttt{FAR} testing splits.
    
    \textbf{Anomaly detection models} We have considered several unsupervised baselines, ranging from classic approaches, like Isolation Forest~\cite{isoforest}, OC-SVM~\cite{ocsvm} and LocalOutlierFactor~\cite{lof}, to deep learning based ones, like the deepSVDD~\cite{deepsvdd} and our proposed transformer for anomalies model, based on the BERT~\cite{bert} architecture.

    \textbf{BERT for anomalies} We employ a simplified BERT architecture, without pretraining, with around 340k trainable parameters. We train the BERT model as a Masked Language Model (MLM), using a data collator that randomly masks $p\%$ of the input sequence by optimizing a cross-entropy loss function between the model predictions at mask positions and the original tokens. We derive a sequence anomaly score by randomly masking $p\%$ of tokens in the sequence and averaging the probabilities of the correct tokens at mask positions given by the classification layer over the vocabulary. At evaluation time, we average the score over $10$ masking samplings. A detailed description of the model is introduced in Appendix~\ref{appendix}. In our experiments, we used $p=15\%$.
    
    In Table~\ref{tab:far_near} we report the results of our experiments. Each baseline model was trained 3 times with a basic set of hyperparameters, and we reported the average results and the standard deviation. The OC-SVM model is solely trained on 10k samples from each yearly split of the \texttt{TRAIN} set to reduce the computational burden. For all of the considered models, we observe a performance degradation between \texttt{NEAR} and \texttt{FAR} splits, highlighting that these anomaly detection models cannot cope with the distribution shift. The \texttt{IID} evaluation, which is the most popular methodology, proves to give an illusion of high performance, as the performance quickly degrades once we consider testing set from a different period. The evolution is also presented in Appendix~\ref{appendix}-Fig.~\ref{fig:perf_gap}, illustrating ROC-AUC along with PR-AUC for inliers and outliers. We observe a rapid degradation for inliers PR-AUC, indicating that normal data distribution is continuously changing, and the outliers detection may not be reliable. These experiments highlight the issues of current anomaly detection models and prove the benefits of the \textbf{AnoShift} benchmark. 
    
\begingroup
\setlength{\tabcolsep}{4pt} % Default value: 6pt
\begin{table}[t]
\begin{center}
    \caption{Performance evolution over time: \texttt{IID} vs \texttt{NEAR} vs \texttt{FAR}. Notice that the ROC-AUC is dropping over time in all cases (small exception for BERT, where the PR-AUC for inliers is also dropping). The variance for \texttt{FAR} is the highest. BERT is the best model for generalizing to \texttt{NEAR}, but interestingly, OC-SVM is better on farther data and LOF on \texttt{IID} split. Best scores per split are shown in bold. PR-AUC for inliers and outliers are available in Appendix~\ref{appendix}-Fig.~\ref{fig:perf_gap} and in Tab.~\ref{aptab:far_near_roc_pr}.}
        \begin{tabular}{l c c c c c}
        
        \toprule
        & \multicolumn{5}{c}{ROC-AUC $\uparrow$}\\
        \cmidrule{2-6}
        \textbf{Split} & \textbf{IsoForest}~\cite{isoforest} & \textbf{OC-SVM}~\cite{ocsvm} & \textbf{deepSVDD}~\cite{deepsvdd} &   \textbf{LOF}~\cite{lof}  & \textbf{BERT~\cite{bert} for anomalies} \\
        \midrule
        
        \texttt{IID} & 78.73 $\pm$ \small 1.23& 76.78 $\pm$ \small 0.43 & 77.87  $\pm$ \small 2.69  & \textbf{86.58} $\pm$ \small 2.35 & 84.54 $\pm$  \small 0.07  \\
        
        \texttt{NEAR}  & 58.08 $\pm$ \small 6.53 & 72.73 $\pm$ \small 	2.02 & 74.78  $\pm$ \small 6.29   & 74.45 $\pm$ \small 1.24  & \textbf{86.05} $\pm$  \small 0.25\\
        
        \texttt{FAR} & 26.54 $\pm$ \small 2.65 & \textbf{46.96} $\pm$ \small 3.07 & 44.33  $\pm$ \small 6.72 & 32.74 $\pm$ \small 12.55  & 28.15 $\pm$  \small 0.06\\
        \bottomrule
        
        \end{tabular}
        \label{tab:far_near}
\end{center}
\end{table}
\endgroup

    \textbf{Performance on \texttt{FAR}} With all tested baselines, we notice a significant decrease in performance for 2014-2015 years for inliers, showing that there might be a particularity with those years. We observe a large distance in the Jeffreys divergence between 2014-2015 and the rest of the years for 2 features: service type and the number of bytes sent by the source IP (see Fig.~\ref{fig:jeff_kyoto}). From the OTDD analysis in Fig.~\ref{fig:otdd_kyoto}, we observe that: first, the inliers from \texttt{FAR} are very distanced to training years; and second, the outliers from \texttt{FAR} are quite close to the training inliers. One root cause of those events can be the steep increase of the "DNS" traffic percentage (from 4\% to 37\%, in 2013, and 2014 respectively). All those reasons describe the shift on \texttt{FAR}, explaining the low performance on it. % Aici era frumos sa mai facem, dar lasam pe rebuttal :( for the inliers class, samples that are miss-classified as we experimentally tested?. \answerTODO{experiment sa demonstreze explicit asta, @marvis: care e acuratetea in 2014 pe filtrul feat1 == DNS; poate insistat pe asta, la o rubrica: unde greseste modelul!!!}. 

\begin{figure}[t]
    \begin{center}
        \includegraphics[width=1\textwidth]{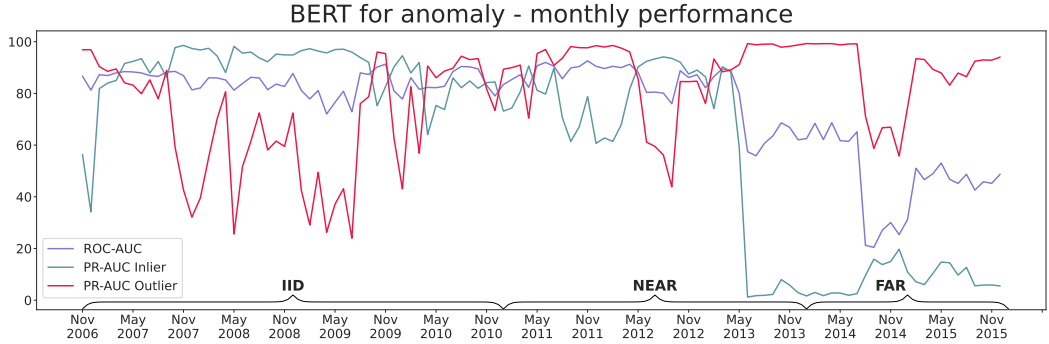}
    \end{center}
    \caption{BERT for anomaly, evaluated on each month. We show the ROC-AUC, PR-AUC for inliers, and PR-AUC for outliers. The performance for the inliers is slowly decreasing  during \texttt{IID} and \texttt{NEAR} splits, dropping suddenly just before the \texttt{FAR} split, showing how the language model fails to recognize inliers once it moves further apart from the training data. On the other hand, there are parts of the \texttt{IID} split where the outliers are quite poorly modeled, explaining the slightly poor performance of BERT on \texttt{IID} when compared with \texttt{NEAR} split.}
    \label{fig:bert_per_month}
    \vspace{-1em}
\end{figure}

    \textbf{Monthly evaluation} In Fig.~\ref{fig:bert_per_month}, we take a closer look at the BERT's performance at month granularity and break down inliers and outliers. First, notice how the inliers' performance gradually degrades over time, to an abrupt drop at farther months. This doubles the analysis from Sec.~\ref{sec:distrib-shift_metrics}, where we notice the difference between the \texttt{TRAIN} years and \texttt{FAR} (through Jeffreys and OTDD experiments). Second, we observe that on \texttt{IID} years, the anomalies are modeled quite poorly by our language model, resulting in a \texttt{IID} performance slightly lower in comparison with \texttt{NEAR}.
    
\subsection{Addressing the shifted data}\label{sec:ack_shift}
\begin{figure}[t]
        \begin{center}
            \includegraphics[width=1\textwidth]{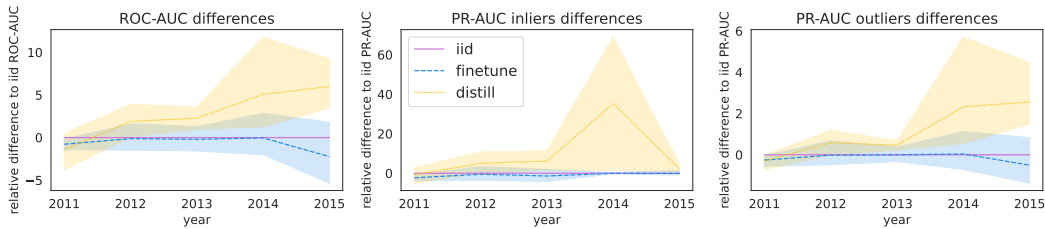}
        \end{center}
        \caption{ROC-AUC, PR-AUC-in, PR-AUC-out for Finetune and Distill strategies, relative to the iid. The performance is averaged over all training subsets. Even though the strategies have a high variance in general, the distill is clearly more robust over time when compared to iid and finetune.}
        \label{fig:iid_finetune_distill}
        \vspace{-1em}
    \end{figure}

     We next compare the performance of a BERT model in 3 training regimes: iid, finetune, and knowledge distillation, for subsets of 300k entries from each year. We use 2006-2010 as training data and evaluate 2011-2015 as individual splits. First, in the \textbf{a) iid mode}, we use sets of data starting from 2006 and gradually add each successive split from the train period, initializing a new model for each subset. Next, in the \textbf{b) finetune mode}, we start from the iid model trained on 2006 and gradually finetune it on each successive year in the train period. Finally, in the \textbf{c) distillation mode}, we start from the iid model of 2006 and reinitialize a same-sized model at each new split, which becomes a student for the previous model by combining the MLM loss with a KL divergence loss with the teacher predictions on the current split. The best performance is achieved by the final distilled model for every test split (see Tab.~\ref{fig:iid_finetune_distill}), outperforming iid and finetune by over $3\%$ on average in ROC-AUC. It is worth noting that the effects of distillation are visible over time, with the iid method outperforming it in the first two iterations over the train splits. At all stages, the distillation method obtains the best performance on FAR data, providing a more robust training alternative to distribution shifts in data. See the exact scores in Appendix~\ref{appendix}-Tab.~\ref{tab:training_strategies}.

% Jeff-features in fig 6.
% 1. Service: the connection's service type, e.g., http, telnet, etc
% 2. Source bytes: the number of data bytes sent by the source IP address 
% 3. Destination bytes: the number of data bytes sent by the destination IP    

% \section{Conclusions}
% \item \answerTODO hypothesis
% \item \answerTODO major findings
% \item \answerTODO contribution 
\vspace{-0.5em}
\section{Conclusion}
Our approach highlights the true dimension of distribution shifts that appear in naturally and continuously evolving data streams. We analyze it in Kyoto-2006\texttt{+} network traffic dataset that spans over 10 years from multiple angles: visually with t-SNE, statistically with histogram distances, and by measuring its magnitude with an Optimal Transport approach. Next, we propose \textbf{AnoShift}, a chronology-based benchmark for anomaly detection, to enable the development of models that generalize better and are more robust to shifts in data. Further, we show that by acknowledging the shift and addressing it, the performance can be improved, obtaining a +3\% performance boost using a basic distillation technique.

\bibliographystyle{plainnat}
\bibliography{neurips_benchmark}

\begin{thebibliography}{49}
\providecommand{\natexlab}[1]{#1}
\providecommand{\url}[1]{\texttt{#1}}
\expandafter\ifx\csname urlstyle\endcsname\relax
  \providecommand{\doi}[1]{doi: #1}\else
  \providecommand{\doi}{doi: \begingroup \urlstyle{rm}\Url}\fi

\bibitem[Aggarwal(2017)]{autoencoder}
Charu~C Aggarwal.
\newblock An introduction to outlier analysis.
\newblock In \emph{Outlier analysis}, pages 1--34. Springer, 2017.

\bibitem[Ahmed et~al.(2016)Ahmed, Mahmood, and Hu]{nids_survey_3}
Mohiuddin Ahmed, Abdun~Naser Mahmood, and Jiankun Hu.
\newblock A survey of network anomaly detection techniques.
\newblock \emph{J. Netw. Comput. Appl.}, 2016.

\bibitem[Alvarez{-}Melis and Fusi(2020)]{otdd}
David Alvarez{-}Melis and Nicol{\`{o}} Fusi.
\newblock Geometric dataset distances via optimal transport.
\newblock In \emph{{NeurIPS}}, 2020.

\bibitem[Arik and Pfister(2021)]{arik2021tabnet}
Sercan~{\"O} Arik and Tomas Pfister.
\newblock Tabnet: Attentive interpretable tabular learning.
\newblock In \emph{Proceedings of the AAAI Conference on Artificial
  Intelligence}, volume~35, pages 6679--6687, 2021.

\bibitem[Breunig et~al.(2000)Breunig, Kriegel, Ng, and Sander]{lof}
Markus~M. Breunig, Hans{-}Peter Kriegel, Raymond~T. Ng, and J{\"{o}}rg Sander.
\newblock {LOF:} identifying density-based local outliers.
\newblock In \emph{{SIGMOD} International Conference on Management of Data},
  2000.

\bibitem[Cai et~al.(2021)Cai, Sener, and Koltun]{intel_cl_benck}
Zhipeng Cai, Ozan Sener, and Vladlen Koltun.
\newblock Online continual learning with natural distribution shifts: An
  empirical study with visual data.
\newblock In \emph{{IEEE/CVF} International Conference on Computer Vision,
  {ICCV}}, 2021.

\bibitem[Chen et~al.(2021)Chen, Weng, Peng, Shuai, and Cheng]{zyell_nctu}
Lei Chen, Shao{-}En Weng, Chu{-}Jun Peng, Hong{-}Han Shuai, and Wen{-}Huang
  Cheng.
\newblock {ZYELL-NCTU} nettraffic-1.0: {A} large-scale dataset for real-world
  network anomaly detection.
\newblock In \emph{{IEEE} International Conference on Consumer Electronics
  {ICCE-TW}}, 2021.

\bibitem[Chen et~al.(2015)Chen, He, Benesty, Khotilovich, Tang, Cho, Chen,
  et~al.]{chen2015xgboost}
Tianqi Chen, Tong He, Michael Benesty, Vadim Khotilovich, Yuan Tang, Hyunsu
  Cho, Kailong Chen, et~al.
\newblock Xgboost: extreme gradient boosting.
\newblock \emph{R package version 0.4-2}, 1\penalty0 (4):\penalty0 1--4, 2015.

\bibitem[Cortes and Vapnik(1995)]{cortes1995support}
Corinna Cortes and Vladimir Vapnik.
\newblock Support-vector networks.
\newblock \emph{Machine learning}, 20\penalty0 (3):\penalty0 273--297, 1995.

\bibitem[Devlin et~al.(2018)Devlin, Chang, Lee, and Toutanova]{devlin2018bert}
Jacob Devlin, Ming-Wei Chang, Kenton Lee, and Kristina Toutanova.
\newblock Bert: Pre-training of deep bidirectional transformers for language
  understanding.
\newblock \emph{arXiv preprint arXiv:1810.04805}, 2018.

\bibitem[Devlin et~al.(2019)Devlin, Chang, Lee, and Toutanova]{bert}
Jacob Devlin, Ming{-}Wei Chang, Kenton Lee, and Kristina Toutanova.
\newblock {BERT:} pre-training of deep bidirectional transformers for language
  understanding.
\newblock \emph{{NAACL}}, 2019.

\bibitem[Dua and Graff(2007)]{kddcup}
Dheeru Dua and Casey Graff.
\newblock Kdd-cup 1999, {UCI} machine learning repository, 2007.
\newblock URL \url{http://kdd.ics.uci.edu/databases/kddcup99/kddcup99.html}.

\bibitem[Gamal et~al.(2020)Gamal, Abbas, and Sadek]{nids_images_svm}
Merna Gamal, Hala Abbas, and Rowayda~A. Sadek.
\newblock Hybrid approach for improving intrusion detection based on deep
  learning and machine learning techniques.
\newblock In \emph{International Conference on Artificial Intelligence and
  Computer Vision, {AICV}}, 2020.

\bibitem[Goodge et~al.(2022)Goodge, Hooi, Ng, and Ng]{lunar}
Adam Goodge, Bryan Hooi, See-Kiong Ng, and Wee~Siong Ng.
\newblock Lunar: Unifying local outlier detection methods via graph neural
  networks.
\newblock In \emph{Proceedings of the AAAI Conference on Artificial
  Intelligence}, volume~36, pages 6737--6745, 2022.

\bibitem[Jeffreys(1946)]{jeff_divergence}
Harold Jeffreys.
\newblock An invariant form for the prior probability in estimation problems.
\newblock \emph{Proc. R. Soc. Lond.}, 1946.

\bibitem[Jr. et~al.(2019)Jr., Rodrigues, Carvalho, Al{-}Muhtadi, and
  Jr.]{nids-survey}
Gilberto~Fernandes Jr., Joel J. P.~C. Rodrigues, Luiz~Fernando Carvalho, Jalal
  Al{-}Muhtadi, and Mario Lemes~Proen{\c{c}}a Jr.
\newblock A comprehensive survey on network anomaly detection.
\newblock \emph{Telecommun. Syst.}, 2019.

\bibitem[K et~al.(2018)K, Vinayakumar, Soman, and
  Poornachandran]{shallow_deep_nids}
Rahul{-}Vigneswaran K, R.~Vinayakumar, K.~P. Soman, and Prabaharan
  Poornachandran.
\newblock Evaluating shallow and deep neural networks for network intrusion
  detection systems in cyber security.
\newblock In \emph{International Conference on Computing, Communication and
  Networking Technologies, {ICCCNT}}, 2018.

\bibitem[Kent(2016)]{lanl}
Alexander~D. Kent.
\newblock Cyber security data sources for dynamic network research.
\newblock In \emph{Dynamic Networks and Cyber-Security}, 2016.

\bibitem[Khraisat et~al.(2019)Khraisat, Gondal, Vamplew, and
  Kamruzzaman]{nids_survey_4}
Ansam Khraisat, Iqbal Gondal, Peter Vamplew, and Joarder Kamruzzaman.
\newblock Survey of intrusion detection systems: techniques, datasets and
  challenges.
\newblock \emph{Cybersecur.}, 2019.

\bibitem[Koh et~al.(2021)Koh, Sagawa, Marklund, Xie, Zhang, Balsubramani, Hu,
  Yasunaga, Phillips, Gao, Lee, David, Stavness, Guo, Earnshaw, Haque, Beery,
  Leskovec, Kundaje, Pierson, Levine, Finn, and Liang]{wilds}
Pang~Wei Koh, Shiori Sagawa, Henrik Marklund, Sang~Michael Xie, Marvin Zhang,
  Akshay Balsubramani, Weihua Hu, Michihiro Yasunaga, Richard~Lanas Phillips,
  Irena Gao, Tony Lee, Etienne David, Ian Stavness, Wei Guo, Berton Earnshaw,
  Imran Haque, Sara~M. Beery, Jure Leskovec, Anshul Kundaje, Emma Pierson,
  Sergey Levine, Chelsea Finn, and Percy Liang.
\newblock {WILDS:} {A} benchmark of in-the-wild distribution shifts.
\newblock In \emph{Proceedings of the International Conference on Machine
  Learning, {ICML}}, 2021.

\bibitem[Kullback and Leibler(1951)]{kl_divergence}
Solomon Kullback and Richard~A Leibler.
\newblock On information and sufficiency.
\newblock \emph{The annals of mathematical statistics}, 1951.

\bibitem[Lazaridou et~al.(2021)Lazaridou, Kuncoro, Gribovskaya, Agrawal, Liska,
  Terzi, Gimenez, de~Masson~d'Autume, Kocisk{\'{y}}, Ruder, Yogatama, Cao,
  Young, and Blunsom]{mind_the_gap}
Angeliki Lazaridou, Adhiguna Kuncoro, Elena Gribovskaya, Devang Agrawal, Adam
  Liska, Tayfun Terzi, Mai Gimenez, Cyprien de~Masson~d'Autume, Tom{\'{a}}s
  Kocisk{\'{y}}, Sebastian Ruder, Dani Yogatama, Kris Cao, Susannah Young, and
  Phil Blunsom.
\newblock Mind the gap: Assessing temporal generalization in neural language
  models.
\newblock In \emph{Advances in Neural Information Processing Systems}, 2021.

\bibitem[Li et~al.(2020)Li, Zhao, Botta, Ionescu, and Hu]{copod}
Zheng Li, Yue Zhao, Nicola Botta, Cezar Ionescu, and Xiyang Hu.
\newblock Copod: copula-based outlier detection.
\newblock In \emph{2020 IEEE International Conference on Data Mining (ICDM)},
  pages 1118--1123. IEEE, 2020.

\bibitem[Li et~al.(2022)Li, Zhao, Hu, Botta, Ionescu, and Chen]{ecod}
Zheng Li, Yue Zhao, Xiyang Hu, Nicola Botta, Cezar Ionescu, and George Chen.
\newblock Ecod: Unsupervised outlier detection using empirical cumulative
  distribution functions.
\newblock \emph{IEEE Transactions on Knowledge and Data Engineering}, 2022.

\bibitem[Liaw et~al.(2002)Liaw, Wiener, et~al.]{liaw2002classification}
Andy Liaw, Matthew Wiener, et~al.
\newblock Classification and regression by randomforest.
\newblock \emph{R news}, 2\penalty0 (3):\penalty0 18--22, 2002.

\bibitem[Lin et~al.(2021)Lin, Shi, Pathak, and Ramanan]{clear}
Zhiqiu Lin, Jia Shi, Deepak Pathak, and Deva Ramanan.
\newblock The {CLEAR} benchmark: Continual learning on real-world imagery.
\newblock In \emph{NeurIPS Datasets and Benchmarks}, 2021.

\bibitem[Liu et~al.(2012)Liu, Ting, and Zhou]{isoforest}
Fei~Tony Liu, Kai~Ming Ting, and Zhi{-}Hua Zhou.
\newblock Isolation-based anomaly detection.
\newblock \emph{{ACM} Trans. Knowl. Discov. Data}, 2012.

\bibitem[Liu et~al.(2019)Liu, Li, Zhou, Jiang, Sun, Wang, and He]{mo_gaal}
Yezheng Liu, Zhe Li, Chong Zhou, Yuanchun Jiang, Jianshan Sun, Meng Wang, and
  Xiangnan He.
\newblock Generative adversarial active learning for unsupervised outlier
  detection.
\newblock \emph{IEEE Transactions on Knowledge and Data Engineering},
  32\penalty0 (8):\penalty0 1517--1528, 2019.

\bibitem[Lo et~al.(2021)Lo, Layeghy, Sarhan, Gallagher, and Portmann]{gnn_nids}
Wai~Weng Lo, Siamak Layeghy, Mohanad Sarhan, Marcus~R. Gallagher, and Marius
  Portmann.
\newblock E-graphsage: A graph neural network based intrusion detection system.
\newblock \emph{ArXiv}, 2021.

\bibitem[Meng et~al.(2019)Meng, Liu, Zhu, Zhang, Pei, Liu, Chen, Zhang, Tao,
  Sun, and Zhou]{loganomaly}
Weibin Meng, Ying Liu, Yichen Zhu, Shenglin Zhang, Dan Pei, Yuqing Liu, Yihao
  Chen, Ruizhi Zhang, Shimin Tao, Pei Sun, and Rong Zhou.
\newblock Loganomaly: Unsupervised detection of sequential and quantitative
  anomalies in unstructured logs.
\newblock In \emph{Artificial Intelligence, {IJCAI}}, 2019.

\bibitem[Mirsky et~al.(2018)Mirsky, Doitshman, Elovici, and
  Shabtai]{ensemble_autoenc_nids}
Yisroel Mirsky, Tomer Doitshman, Yuval Elovici, and Asaf Shabtai.
\newblock Kitsune: An ensemble of autoencoders for online network intrusion
  detection.
\newblock In \emph{Network and Distributed System Security Symposium, {NDSS}}.
  The Internet Society, 2018.

\bibitem[Moustafa and Slay(2015)]{unsw_nb15}
Nour Moustafa and Jill Slay.
\newblock {UNSW-NB15:} a comprehensive data set for network intrusion detection
  systems {(UNSW-NB15} network data set).
\newblock In \emph{Military Communications and Information Systems Conference,
  MilCIS}, 2015.

\bibitem[Pang et~al.(2021)Pang, Shen, Cao, and van~den Hengel]{DL_anom_survey}
Guansong Pang, Chunhua Shen, Longbing Cao, and Anton van~den Hengel.
\newblock Deep learning for anomaly detection: {A} review.
\newblock \emph{{ACM} Comput. Surv.}, 2021.

\bibitem[P{\'{e}}rez et~al.(2019)P{\'{e}}rez, Alonso, {\'{A}}lvarez, Prada,
  Fuertes, and Dom{\'{\i}}nguez]{compare_nids}
Daniel P{\'{e}}rez, Seraf{\'{\i}}n Alonso, Antonio~Mor{\'{a}}n {\'{A}}lvarez,
  Miguel~A. Prada, Juan~Jos{\'{e}} Fuertes, and Manuel Dom{\'{\i}}nguez.
\newblock Comparison of network intrusion detection performance using feature
  representation.
\newblock In \emph{Engineering Applications of Neural Networks{EANN}}, 2019.

\bibitem[Ring et~al.(2019)Ring, Wunderlich, Scheuring, Landes, and
  Hotho]{ring2019survey}
Markus Ring, Sarah Wunderlich, Deniz Scheuring, Dieter Landes, and Andreas
  Hotho.
\newblock A survey of network-based intrusion detection data sets.
\newblock \emph{Computers \& Security}, 86:\penalty0 147--167, 2019.

\bibitem[Ruff et~al.(2018)Ruff, Vandermeulen, G{\"o}rnitz, Deecke, Siddiqui,
  Binder, M{\"u}ller, and Kloft]{deepsvdd}
Lukas Ruff, Robert~A. Vandermeulen, Nico G{\"o}rnitz, Lucas Deecke, Shoaib~A.
  Siddiqui, Alexander Binder, Emmanuel M{\"u}ller, and Marius Kloft.
\newblock Deep one-class classification.
\newblock In \emph{International Conference on Machine Learning {ICML}}, 2018.

\bibitem[Sarhan et~al.(2020)Sarhan, Layeghy, Moustafa, and
  Portmann]{netflow_v1_iot}
Mohanad Sarhan, Siamak Layeghy, Nour Moustafa, and Marius Portmann.
\newblock Netflow datasets for machine learning-based network intrusion
  detection systems.
\newblock In \emph{Big Data Technologies and Applications Conference, {BDTA},
  and International Conference on Wireless Internet, WiCON}, 2020.

\bibitem[Sarhan et~al.(2022)Sarhan, Layeghy, and Portmann]{netflow_v2_iot}
Mohanad Sarhan, Siamak Layeghy, and Marius Portmann.
\newblock Towards a standard feature set for network intrusion detection system
  datasets.
\newblock \emph{Mob. Networks Appl.}, 2022.

\bibitem[Sch{\"{o}}lkopf et~al.(1999)Sch{\"{o}}lkopf, Williamson, Smola,
  Shawe{-}Taylor, and Platt]{ocsvm}
Bernhard Sch{\"{o}}lkopf, Robert~C. Williamson, Alexander~J. Smola, John
  Shawe{-}Taylor, and John~C. Platt.
\newblock Support vector method for novelty detection.
\newblock In \emph{Advances in Neural Information Processing Systems, {NIPS}},
  1999.

\bibitem[Sharafaldin et~al.(2018)Sharafaldin, Lashkari, and
  Ghorbani]{cic_ids_2017}
Iman Sharafaldin, Arash~Habibi Lashkari, and Ali~A. Ghorbani.
\newblock Toward generating a new intrusion detection dataset and intrusion
  traffic characterization.
\newblock In \emph{International Conference on Information Systems Security and
  Privacy, {ICISSP}}, 2018.

\bibitem[Shenkar and Wolf(2021)]{icl}
Tom Shenkar and Lior Wolf.
\newblock Anomaly detection for tabular data with internal contrastive
  learning.
\newblock In \emph{International Conference on Learning Representations}, 2021.

\bibitem[Somepalli et~al.(2021)Somepalli, Goldblum, Schwarzschild, Bruss, and
  Goldstein]{somepalli2021saint}
Gowthami Somepalli, Micah Goldblum, Avi Schwarzschild, C~Bayan Bruss, and Tom
  Goldstein.
\newblock Saint: Improved neural networks for tabular data via row attention
  and contrastive pre-training.
\newblock \emph{arXiv preprint arXiv:2106.01342}, 2021.

\bibitem[Song et~al.(2011)Song, Takakura, Okabe, Eto, Inoue, and
  Nakao]{kyoto2006}
Jungsuk Song, Hiroki Takakura, Yasuo Okabe, Masashi Eto, Daisuke Inoue, and
  Koji Nakao.
\newblock Statistical analysis of honeypot data and building of kyoto 2006+
  dataset for {NIDS} evaluation.
\newblock In \emph{Proceedings of the First Workshop on Building Analysis
  Datasets and Gathering Experience Returns for Security, BADGERS EuroSys},
  2011.

\bibitem[Sun et~al.(2022)Sun, Segu, Postels, Wang, Van~Gool, Schiele, Tombari,
  and Yu]{shift_auto_drive}
Tao Sun, Mattia Segu, Janis Postels, Yuxuan Wang, Luc Van~Gool, Bernt Schiele,
  Federico Tombari, and Fisher Yu.
\newblock {SHIFT:} a synthetic driving dataset for continuous multi-task domain
  adaptation.
\newblock In \emph{Computer Vision and Pattern Recognition, {CVPR}}, 2022.

\bibitem[Tavallaee et~al.(2009)Tavallaee, Bagheri, Lu, and Ghorbani]{NSL-KDD}
Mahbod Tavallaee, Ebrahim Bagheri, Wei Lu, and Ali~A. Ghorbani.
\newblock A detailed analysis of the {KDD} {CUP} 99 data set.
\newblock In \emph{Symposium on Computational Intelligence for Security and
  Defense Applications, {CISDA}}, 2009.

\bibitem[van~der Maaten and Hinton(2008)]{tsne}
Laurens van~der Maaten and Geoffrey Hinton.
\newblock Visualizing data using {t-SNE}.
\newblock \emph{Journal of Machine Learning Research}, 2008.

\bibitem[Verkerken et~al.(2022)Verkerken, D'hooge, Wauters, Volckaert, and
  Turck]{nids_generalization}
Miel Verkerken, Laurens D'hooge, Tim Wauters, Bruno Volckaert, and Filip~De
  Turck.
\newblock Towards model generalization for intrusion detection: Unsupervised
  machine learning techniques.
\newblock \emph{J. Netw. Syst. Manag.}, 2022.

\bibitem[Yang et~al.(2019)Yang, Moubayed, Hamieh, and Shami]{tree}
Li~Yang, Abdallah Moubayed, Ismail Hamieh, and Abdallah Shami.
\newblock Tree-based intelligent intrusion detection system in internet of
  vehicles.
\newblock In \emph{{IEEE} Global Communications Conference, {GLOBECOM}}, 2019.

\bibitem[Zhao et~al.(2019)Zhao, Nasrullah, and Li]{pyod}
Yue Zhao, Zain Nasrullah, and Zheng Li.
\newblock Pyod: A python toolbox for scalable outlier detection.
\newblock \emph{Journal of Machine Learning Research}, 20\penalty0
  (96):\penalty0 1--7, 2019.
\newblock URL \url{http://jmlr.org/papers/v20/19-011.html}.

\end{thebibliography}

%%%%%%%%%%%%%%%%%%%%%%%%%%%%%%%%%%%%%%%%%%%%%%%%%%%%%%%%%%%%
\clearpage
\section*{Checklist}

% %%% BEGIN INSTRUCTIONS %%%
% The checklist follows the references.  Please
% read the checklist guidelines carefully for information on how to answer these
% questions.  For each question, change the default \answerTODO{} to \answerYes{}, \answerNo{}, or \answerNA{}.  You are strongly encouraged to include a {\bf justification to your answer}, either by referencing the appropriate section of
% your paper or providing a brief inline description.  For example:
% \begin{itemize}
%   \item Did you include the license to the code and datasets? \answerYes{See Section~X.}
%   \item Did you include the license to the code and datasets? \answerNo{The code and the data are proprietary.}
%   \item Did you include the license to the code and datasets? \answerNA{}
% \end{itemize}
% Please do not modify the questions and only use the provided macros for your
% answers.  Note that the Checklist section does not count towards the page
% limit.  In your paper, please delete this instructions block and only keep the
% Checklist section heading above along with the questions/answers below.
% %%% END INSTRUCTIONS %%%

\begin{enumerate}

\item For all authors...
\begin{enumerate}
  \item Do the main claims made in the abstract and introduction accurately reflect the paper's contributions and scope? 
    \answerYes{}
  \item Did you describe the limitations of your work?
    \answerYes{See Sec.~\ref{sec:kyoto} - paragraph 2.; Sec. \ref{sec:exp_setup} - paragraph 2.; Appendix~\ref{appendix:discussions}}
  \item Did you discuss any potential negative societal impacts of your work?
  
    \answerNo{Our work does not have a negative societal impact. Our benchmark proposal is tailored for finding intrusions in a computer network (not at the user level, but at the network level), by detecting anomalous traffic, in a more robust way than before, closer to the real scenario. One use-case is in the IT department of an company or university, where a person monitors the traffic alerts and prioritizes certain alerts based on the predictions of the robust models trained on our proposed benchmark.}
  \item Have you read the ethics review guidelines and ensured that your paper conforms to them?
    \answerYes{}
\end{enumerate}

\item If you are including theoretical results...
\begin{enumerate}
  \item Did you state the full set of assumptions of all theoretical results?
    \answerNA{}
	\item Did you include complete proofs of all theoretical results?
    \answerNA{}
\end{enumerate}

\item If you ran experiments (e.g. for benchmarks)...
\begin{enumerate}
  \item Did you include the code, data, and instructions needed to reproduce the main experimental results (either in the supplemental material or as a URL)?
    \answerYes{in the abstract and the appendix}.
  \item Did you specify all the training details (e.g., data splits, hyperparameters, how they were chosen)?
    \answerYes{see Sec.~\ref{sec:distrib_shift_analysis} and its subsections}.
	\item Did you report error bars (e.g., with respect to the random seed after running experiments multiple times)?
    \answerYes{For each of the tested methods in the main experiment, we run it 3 times, with different seeds.} 
	\item Did you include the total amount of compute and the type of resources used (e.g., type of GPUs, internal cluster, or cloud provider)?
    \answerYes{see Sec.~\ref{sec:distrib_shift_analysis} - paragraph 2} 

\end{enumerate}

\item If you are using existing assets (e.g., code, data, models) or curating/releasing new assets...
\begin{enumerate}
  \item If your work uses existing assets, did you cite the creators?
    \answerYes{We used the existing Kyoto-2006\texttt{+} dataset as raw data, as detailed in Sec.~\ref{sec:chronological_protocol}}
  \item Did you mention the license of the assets?
    \answerNA{The authors do not mention any kind of licence for the data, it is just publicly available.}
  \item Did you include any new assets either in the supplemental material or as a URL?
    \answerYes{We included a GitHub repository with code resources and a repository with the preprocessed data in the suplimentary material - see Appendix.~\ref{sec:appendix_own}}
  \item Did you discuss whether and how consent was obtained from people whose data you're using/curating?
    \answerNo{We had several emails with the authors, describing them our purpose and asking for additional information.}
  \item Did you discuss whether the data you are using/curating contains personally identifiable information or offensive content?
    \answerYes{see in Sec.~\ref{sec:exp_setup}}

\end{enumerate}

\item If you used crowdsourcing or conducted research with human subjects...
\begin{enumerate}
  \item Did you include the full text of instructions given to participants and screenshots, if applicable?
    \answerNA{}
  \item Did you describe any potential participant risks, with links to Institutional Review Board (IRB) approvals, if applicable?
    \answerNA{}
  \item Did you include the estimated hourly wage paid to participants and the total amount spent on participant compensation?
    \answerNA{}
\end{enumerate}

\end{enumerate}

%%%%%%%%%%%%%%%%%%%%%%%%%%%%%%%%%%%%%%%%%%%%%%%%%%%%%%%%%%%%
\clearpage
\appendix

\section{Appendix}\label{appendix}

\paragraph{Kyoto-2006\texttt{+}} When an attack occurs, the honeypot saves the network access pattern and other metadata and, at some point, might decide to reboot the system and rewrite back the original configuration. The authors deployed another machine in the network to generate normal traffic data, with a mailing server and a DNS service for a single domain. All traffic data from this server was labeled as clean (the logs also include other protocols for managing the machine over ssh or http and https). The 14 conventional features of the dataset includes 2 categorical features: connection service type and flag of the connection and 12 numerical features: connection duration, number of source and destination bytes, number of connections with corresponding IP addresses in a timeframe of two seconds and the percentage of connections accessing the same service and their rate of "SYN" errors, the prevalence of the connection's source IP address and the requested service in the past 100 connections to the current destination IP (\href{http://www.takakura.com/Kyoto_data/BenchmarkData-Description-New.pdf}{Kyoto features}). The malicious traffic is further labeled using three software solutions: an Intrusion Detection Systems at the network level, an Antivirus product, and a shellcodes and exploits detector. In addition to those, the authors labeled other entries based on their prior history of connections from a specific IP and destination port. Additional features include source and destination IP addresses and ports, timestamp and protocol. We note that the generality of the original dataset imposes several limitations for our benchmark. More specific, the diversity of the normal traffic in a honeypot setup is quite restricted. Also, since the labeling is done using existing software and rules, the dataset's anomalies might be underestimated.

\paragraph{Performance evolution over time} In Tab.~\ref{aptab:far_near_roc_pr} we present the full evaluation of considered baseline models on \texttt{IID}, \texttt{NEAR} and \texttt{FAR} splits. 

\paragraph{Training strategies for data shift} In Tab.~\ref{tab:training_strategies} we present the full ROC-AUC, and PR-AUC for inliers and outliers, for all three training strategies: iid, finetune and distill.

\paragraph{BERT for Anomalies} We propose a simplified BERT architecture for detecting anomalies. The network input is tokenized by a WordLevel tokenizer which obtains tokens for the individual events in a system log sequence and, conversely, for the individual features of network traffic. Therefore, we have fixed-length sequences for Kyoto-2006. We train the BERT model as a Masked Language Model (MLM), using a data collator that randomly masks $p\%$ of the input sequence, by optimizing a cross-entropy loss function between the model predictions at mask positions and the original tokens. We derive a sequence anomaly score by randomly masking $p\%$ of tokens in the sequence and averaging the probabilities of the correct tokens at mask positions given by the classification layer over the vocabulary. The model is not pretrained and consists of two hidden layers of size $120$, an intermediate size of $192$ and $6$ attention heads. It has a hidden dropout and attention dropout probabilities of $0.1$, an epsilon of $1e-12$ for the normalization layer and a $0.02$ standard deviation range for the truncated normal weight initialization. Our architecture totals $342135$ trainable parameters. For training we mask $p=15\%$ of the input sequence and at evaluation time, we average over $n=10$ mask samplings.

\begingroup
\setlength{\tabcolsep}{4pt} % Default value: 6pt
\begin{table}
\begin{center}
    \caption{Performance evolution over time: \texttt{IID} vs \texttt{NEAR} vs \texttt{FAR}. We report beside the ROC-AUC metric, also the PR-AUC for inliers and PR-AUC for outliers.}
        \begin{tabular}{l c c c c c}
        \toprule
        
        & \textbf{IsoForest}~\cite{isoforest} & \textbf{OC-SVM}~\cite{ocsvm} & \textbf{deepSVDD}~\cite{deepsvdd} &   \textbf{LOF}~\cite{lof}  & \textbf{BERT~\cite{bert} for anomalies} \\
        \cmidrule{2-6}
       \textbf{Split} & \multicolumn{5}{c}{ROC-AUC $\uparrow$} \\
        \cmidrule{2-6}
        % \textbf{Split} & \textbf{IsoForest}~\cite{isoforest} & \textbf{OC-SVM}~\cite{ocsvm} & \textbf{deepSVDD}~\cite{deepsvdd} &   \textbf{LOF}~\cite{lof}  & \textbf{BERT~\cite{bert} for anomalies} \\
        % \midrule
        
        \texttt{IID} & 78.73 $\pm$ \small 1.23& 76.78 $\pm$ \small 0.43 & 77.87  $\pm$ \small 2.69  & 86.58 $\pm$ \small 2.35 & 84.54 $\pm$  \small 0.07  \\
        
        \texttt{NEAR}  & 58.08 $\pm$ \small 6.53 & 72.73 $\pm$ \small 	2.02 & 74.78  $\pm$ \small 6.29   & 74.45 $\pm$ \small 1.24  & 86.05 $\pm$  \small 0.25\\
        
        \texttt{FAR} & 26.54 $\pm$ \small 2.65 & 46.96 $\pm$ \small 3.07 & 44.33  $\pm$ \small 6.72 & 32.74 $\pm$ \small 12.55  & 28.15 $\pm$  \small 0.06\\
        
        \cmidrule{2-6}
        &\multicolumn{5}{c}{PR-AUC-inliers $\uparrow$}\\
        \cmidrule{2-6}
        \texttt{IID}  & 70.38 $\pm$ \small  1.09 &  69.46 $\pm$ \small 1.67 &   75.52 $\pm$ \small 0.77 &  78.77 $\pm$ \small 0.67 & 74.61 $\pm$  \small0.13 \\
        
        \texttt{NEAR}   & 29.04 $\pm$ \small 3.50 &  39.75$\pm$ \small 	0.93 &   57.79$\pm$ \small 	7.99 & 44.71 $\pm$ \small 4.65 &  58.94$\pm$  \small 	0.69\\
        
        \texttt{FAR} & 9.15 $\pm$ \small 0.23 & 13.14 $\pm$ \small 	1.98 &  12.18 $\pm$ \small 1.95 & 10.40 $\pm$ \small  2.32 & 8.22 $\pm$  \small 0.02\\

        % \toprule
        \cmidrule{2-6}
        & \multicolumn{5}{c}{PR-AUC-outliers $\uparrow$}\\
        \cmidrule{2-6}
        \texttt{IID}  &81.61 $\pm$ \small 	2.12 &  70.43$\pm$ \small2.96  & 61.67  $\pm$ \small 	3.84	 & 	83.04 $\pm$ \small 4.70 & 	89.83 $\pm$  \small 0.07 \\
        
        \texttt{NEAR}   & 81.44$\pm$ \small 3.58 & 87.24 $\pm$ \small  2.15& 86.00  $\pm$ \small 	3.61 &  89.60	 $\pm$ \small	1.29  &  	95.96 $\pm$  \small 0.06 \\
        
        \texttt{FAR} & 78.70$\pm$ \small 0.75 & 86.80 $\pm$ \small 0.55 & 	84.79  $\pm$ \small 2.04 &  78.72 $\pm$ \small	4.51  &  78.38$\pm$  \small 0.02\\
        \bottomrule
        \end{tabular}
        \label{aptab:far_near_roc_pr}
\end{center}

\end{table}

\begin{equation}
    \hat{w_j}^i =  \begin{cases}
                        w_j, & \text{if $mask_i$(j) = 0} \\
                        [MASK], & \text{if $mask_i$(j) = 1}
                    \end{cases}
    \label{masking}
\end{equation}

We repeat the masking process n times and average over all repeats to improve consistency. The anomaly score formula is depicted in equation \ref{anom_score}, where we denote by $P_M$ the classification layer of the model of parameters $\theta_M$, by $Masks^{p}_{k}$ the set of random binary masks of length k and mask probability p, where $w_t$ are the initial tokens in the sequence and $\hat{w_t}^i$ is the j-th token in the sequence under mask i.

\begin{equation}
anomaly\_score([w_1, w_2, ... , w_t]) = \frac{\sum_{i=1..n} \sum_{j=1..t}^{mask_i \sim Masks^{p}_t} (1 - P(\hat{w_j}^i)) }{n}
\label{anom_score}
\end{equation}

\begin{equation}
    P(\hat{w_j}^i) =  \begin{cases}
                        1, & \text{if $mask_i$(j) = 0} \\
                        P_M(w_j | \theta_M, [\hat{w_1}^i, ..., \hat{w_t}^i]), & \text{if $mask_i$(j) = 1}
                    \end{cases}
    \label{classif_prob}
\end{equation}

We motivate this metric with the observation that inlier data should consist of common tokens with a high retrieval probability given by the distribution of the training set, while outliers should usually have either rare tokens or unusual combinations of features, within the context given by the unmasked tokens.

\begin{table}[t]
\caption{Training strategies: ROC-AUC for IID training vs Finetune vs Distil on Kyoto-2006\texttt{+}}
\setlength{\tabcolsep}{7pt} % Default value: 6pt
\begin{center}
\begin{tabular}{c clllc llllc }

\toprule
 & \shortstack{Train ON: \textbf{2006 ->}} & \multicolumn{1}{c}{\textbf{2007}$\uparrow$} & \multicolumn{1}{c}{\textbf{2008}$\uparrow$} & \multicolumn{1}{c}{\textbf{2009}$\uparrow$} & \multicolumn{1}{c}{\textbf{2010}$\uparrow$} \\ 
 \cmidrule{2-2}
\textbf{Strategy} & \shortstack{Test split} \\
\toprule

\multicolumn{1}{l}{\textbf{IID}} & 2011 & 88.95 & 88.93 & 88.27 & 89.92 \\
& 2012 & 95.85 & 90.96 & 86.28 & 86.63
\\ & 2013 & 94.05 &	87.00 & 80.79 & 81.87 \\ & 2014 & 28.56 & 24.35 & 22.64 & 21.16\\ & 2015 & 49.01 & 42.20 & 37.05 & 34.07\\
\midrule
\multicolumn{1}{l}{\textbf{Finetune}} & 2011 & 88.56 & 87.18 & 87.3 & 89.92\\
& 2012 & 95.78 & 89.31 & 85.18 & 89.14\\ & 2013 & 94.19 & 85.29 & 79.36 & 84.06\\ & 2014 & 32.98 & 22.50 & 20.50 & 20.57\\ & 2015 & 53.39 & 38.99 & 30.98 & 30.04\\
\midrule
\multicolumn{1}{l}{\textbf{Distil}}  & 2011 & 83.74 & 87.43 & 88.82 & 90.32\\
& 2012 & 95.10 & 91.96 & 88.78 & 91.51\\ & 2013 & 94.43 & 88.89 & 83.43 & 86.04\\ & 2014 & 43.65 & 26.69 & 23.55 & 23.13\\ & 2015 & 59.93 & 48.39 & 41.31 & 39.69\\

\bottomrule

\end{tabular}
\label{tab:training_strategies}
\end{center}
\end{table}

\paragraph{Impure training} Till now, we have considered anomaly detection models that were trained solely on the clean data of the \texttt{TRAIN} split. Further, we will study the performance evolution between \texttt{IID}, \text{NEAR} and \texttt{FAR} when the BERT model is trained on corrupt data containing mislabeled outliers in different percentages. In Fig.~\ref{fig:impure} we present the ROC-AUC for the 3 testing splits. We observe that the distribution shift is noticeable in the model performance even in this corrupt training setup. This validates the usefulness of the proposed chronological protocol when dealing with potentially mislabeled samples and highlights that the observed degradation over time is not a consequence of such dataset issues. 

\begin{figure}[t]
    \begin{center}
        \includegraphics[width=0.4\textwidth]{images_/impure.png}
    \end{center}
    \caption{Performance evolution of our BERT model on \texttt{IID}, \texttt{NEAR} and \texttt{FAR} splits, when training on a corrupt set of samples, containing different percentages of mislabeled data points. (Best viewed in color)}
    \label{fig:impure}
\end{figure}

\begin{figure}[t]
    \begin{center}
        \includegraphics[width=1\textwidth]{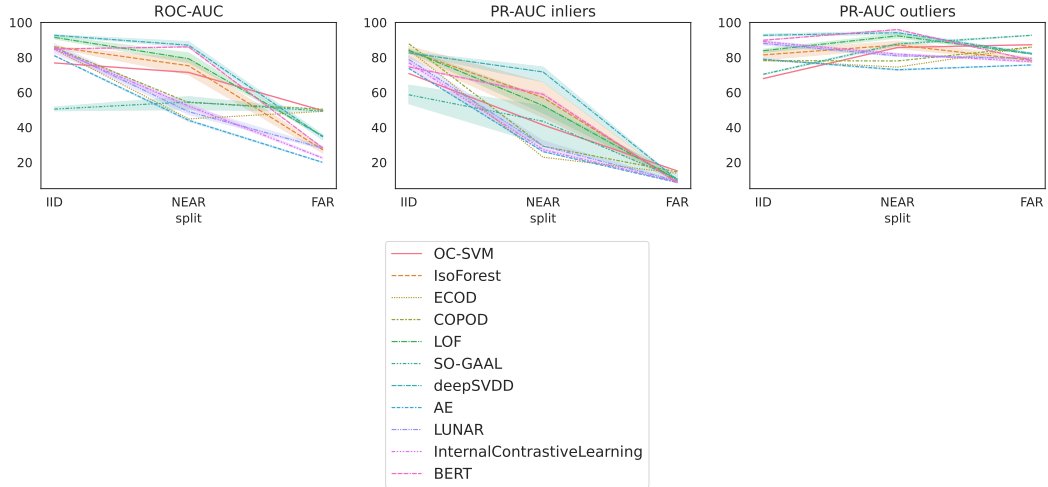}
    \end{center}
    \caption{Performance evolution over time: \texttt{IID} vs \texttt{NEAR} vs \texttt{FAR}.  We follow the evolution of ROC-AUC and PR-AUC for inliers and outliers. We observe a large performance gap between the considered splits, correlated with the temporal distance from the training set. (Best viewed in color)}
    \label{fig:perf_gap}
\end{figure}

\paragraph{Broader Impact} Our benchmark proposal is tailored for finding intrusions in a computer network (not at the user level, but at the network level), by detecting anomalous traffic, in a more robust way than before, closer to the real scenario. One use-case is in the IT department of an company or university, where a person monitors the traffic alerts and prioritizes certain alerts based on the predictions of the robust models trained on our proposed benchmark. Our work does not have a negative societal impact.

\subsection{Discussions and future work}
\label{appendix:discussions}
% \item \answerTODO major findings and their meaning \\
% \item \answerTODO link to other works - w.r.t findings\\
% \item \answerTODO limitations - w.r.t findings\\
% \item \answerTODO explanations for unexpected / inconclusive results \\
% \item \answerTODO limitations
% \item \answerTODO future directions 

% \paragraph{Annotation limitation} The labels used for evaluating the algorithms are composed out of 3 software predictions and several rules based on previous detected intrusions. There are for sure missing 

\paragraph{MLM as anomaly detector} Even though the BERT model greatly exceeds the number of parameters and the complexity of other baselines, its generalization performance on farther data is extremely low. The anomaly performance in our case is based on the perplexity score when predicting several masked features in the sample. So if the features are not correlated, the MLM model might be unable to learn something useful. We did not investigate this, but we consider it an interesting direction for future work.

\paragraph{MLM with the training vocabulary} If the model only learns based on the training data, all the new words (in our case, features from a new sample) will be considered UNK, and the score might get artificially inflated (since from a point further it is easier to predict UNK instead of the right word). This might be an disadvantage of vocabulary based methods as opposed to other classic approaches. Even though this does not apply to our \textbf{AnoShift} benchmark, might be a problem for similar scenarios (\eg system logs datasets).

\paragraph{Multivariate time series} For better capturing the links between network traffic packets, we might need very dense samples, like the continuous stream captured from the network. Even though it seems costly, we plan to analyze and extend \textbf{AnoShift} (if worth it) to cover the full Kyoto-2006\texttt{+} dataset, enabling multivariate time series models over shifted data.

\paragraph{The inliers' natural distribution} For being able to annotate large amount of data, the network datasets stay either in a clean space, where almost everything is normal, or in a "dark" one, where every connection is considered infected. Kyoto-2006\texttt{+} lies in the second case, where the normal traffic is not very general, covering several behaviours. It might be interesting as future directions to find a way to combine the two cases towards a more general and unbiased dataset.

\paragraph{Pre-process through binning} Because we transform the features into vocabulary, we binned the numerical ones, loosing some pieces of information. Depending on the algorithm, this might not be necessarily needed.

\clearpage

\section{Appendix}
\label{sec:appendix_own}
\paragraph{Raw Kyoto dataset documentation}
The dataset used in our proposed benchmark consists of a preprocessing of the Traffic Data from Kyoto University's Honeypots and results from a discretization of the numerical features in the original dataset, such that a language-modelling approach can be easily applied. The original dataset consists of 14 conventional features and 10 additional features, of which we discard the source and destination IP addresses, source and destination port numbers. The conventional features in the original dataset includes connection duration, type of service, number of source and destination bytes, server rate errors percentage and flag of connection. We keep all the conventional features and apply a exponentially-scaled binning over the continuous values (duration, number of source and destination bytes) which results in 233 bins and a discretization of the percentage features in 100 distinct values.

\paragraph{Our split proposal documentation}
We propose a yearly split of the dataset and group adjacent years into Train, NEAR data and FAR data, which we use to highlight the performance degradation of several benchmarks in time, due to the distributional shift of the data which we demonstrate with a comprehensive analysis. In our proposed split, Train data consists of the first 4 years (2006-2010), Near data of the following 3 years (2011-2013) and Far data of the last two available years (2014-2015). We publish the data in splits of single years, in csv format. The columns 0 to 13 are the discretized conventional features in the Kyoto-2006+ dataset, preserving the original order, column 14 contains the complete timestamp. Columns 15, 16 and 17 correspond to the first 3 additional features in the original data, namely IDS\_detection, Malware\_detection and Ashula\_detection, which indicates presence of alert triggers from the 3 IDS solutions: Symantec IDS, clamav and Ashula shellcode detector. Column 19 in the preprocessed dataset corresponds to the protocol used by the connection.
\paragraph{Intended uses}
We hope that our proposed benchmark shifts the general direction of treating network intrusion detection towards a timely fashion that suffers from distributional shift, hereby providing a better suited evaluation protocol for upcoming research in this field.

\paragraph{URL to dataset download}
We redirect our readers to the repository of the raw Kyoto dataset published by the Kyoto University at \url{https://www.takakura.com/Kyoto_data/} and provide a repository of data under our proposed processing at \url{https://share.bitdefender.com/apps/files/?dir=/Deeplearning/Kyoto-2016_subsets&fileid=716592}, with subsets of 300000 instances and heldout sets of 30000 instances for each split, maintaining the original inlier to outlier ratio from the original data in each split, as well as the full processed splits, with each full split except 2006 being provided in two parts. We make the remark that the 2006 split contains fewer instance, due to data collection debuting in November.

\subsection{Code for dataset loading}
We publish our code as a public GitHub repository \url{https://github.com/bit-ml/AnoShift/}, containing the data preprocessing script that transforms the original data in our format, sample data manipulation notebooks, license and additional information.

\subsection{Author responsibility for violation of rights}
There is no sensitive data leaked in the preprocessed dataset. The authors are not aware of any possible violation of rights and take responsibility for the published data.

\subsection{Dataset hosting and long-term preservation}
The authors take full responsibility for the availability of the processed data in the provided repository. However, no statement can be made about the availability of the raw Kyoto-2006+ data published by the Kyoto University.

\subsection{Licence}
We release our code under a BSD 3-Clause License, therefore allowing the redistribution and use in source and binary forms, with or without
modification, under the 3 clauses specified by the Berkeley Software Distribution License:

\begin{enumerate}
    \item{Redistributions of source code must retain the above copyright notice, this list of conditions and the following disclaimer.
    }
    \item{Redistributions in binary form must reproduce the above copyright notice, this list of conditions and the following disclaimer in the documentation and/or other materials provided with the distribution.}
    
    \item{Neither the name of the copyright holder nor the names of its
    contributors may be used to endorse or promote products derived from
    this software without specific prior written permission.}
\end{enumerate}

\end{document}